\documentclass[11pt,twoside]{article}
\usepackage[latin9]{inputenc}
\usepackage{float}
\usepackage{amsmath}
\usepackage{amsthm}
\usepackage{amssymb}
\usepackage{graphicx}
\usepackage[authoryear]{natbib}
\usepackage{xargs}[2008/03/08]

\makeatletter
\theoremstyle{plain}
\newtheorem{thm}{\protect\theoremname}
  \theoremstyle{plain}
  \newtheorem{prop}[thm]{\protect\propositionname}

\usepackage{jmlr2e}
\usepackage{natbib}
\usepackage{flushend}
\usepackage{graphicx}
\usepackage{algolyx}

\jmlrheading{x}{2017}{y-z}{a/b}{c/d}{Trung Le, Tu Dinh Nguyen,  and Dinh Phung}

\ShortHeadings{Kernelized Generative Adversarial Networks}{Le et al}
\firstpageno{1}

\usepackage{colortbl} 
\definecolor{header_color}{rgb}{0.74,0.88,0.91}
\definecolor{even_color}{rgb}{0.9,0.9,0.9}
\definecolor{subheader_color}{rgb}{0.85,0.93,0.95}
\definecolor{childheader_color}{rgb}{1.0,0.93,0.87}

\definecolor{ccolor_best}{rgb}{1.0,0.9,0.9}
\definecolor{ccolor_wrong}{rgb}{1.0,0.85,0.85}

\newcolumntype{x}[1]{>{\centering\arraybackslash}p{#1}}

\makeatother

  \providecommand{\propositionname}{Proposition}
\providecommand{\theoremname}{Theorem}

\begin{document}
\newcommand{\sidenote}[1]{\marginpar{\small \emph{\color{Medium}#1}}}

\global\long\def\se{\hat{\text{se}}}

\global\long\def\interior{\text{int}}

\global\long\def\boundary{\text{bd}}

\global\long\def\ML{\textsf{ML}}

\global\long\def\GML{\mathsf{GML}}

\global\long\def\HMM{\mathsf{HMM}}

\global\long\def\support{\text{supp}}

\global\long\def\new{\text{*}}

\global\long\def\stir{\text{Stirl}}

\global\long\def\mA{\mathcal{A}}

\global\long\def\mB{\mathcal{B}}

\global\long\def\mF{\mathcal{F}}

\global\long\def\mK{\mathcal{K}}

\global\long\def\mH{\mathcal{H}}

\global\long\def\normal{\mathcal{N}}

\global\long\def\mX{\mathcal{X}}

\global\long\def\mZ{\mathcal{Z}}

\global\long\def\mS{\mathcal{S}}

\global\long\def\Ical{\mathcal{I}}

\global\long\def\mT{\mathcal{T}}

\global\long\def\Pcal{\mathcal{P}}

\global\long\def\dist{d}

\global\long\def\HX{\entro\left(X\right)}
 \global\long\def\entropyX{\HX}

\global\long\def\HY{\entro\left(Y\right)}
 \global\long\def\entropyY{\HY}

\global\long\def\HXY{\entro\left(X,Y\right)}
 \global\long\def\entropyXY{\HXY}

\global\long\def\mutualXY{\mutual\left(X;Y\right)}
 \global\long\def\mutinfoXY{\mutualXY}

\global\long\def\given{\mid}

\global\long\def\gv{\given}

\global\long\def\goto{\rightarrow}

\global\long\def\asgoto{\stackrel{a.s.}{\longrightarrow}}

\global\long\def\pgoto{\stackrel{p}{\longrightarrow}}

\global\long\def\dgoto{\stackrel{d}{\longrightarrow}}

\global\long\def\lik{\mathcal{L}}

\global\long\def\logll{\mathit{l}}

\global\long\def\vectorize#1{\mathbf{#1}}

\global\long\def\vt#1{\mathbf{#1}}

\global\long\def\gvt#1{\boldsymbol{#1}}

\global\long\def\idp{\ \bot\negthickspace\negthickspace\bot\ }
 \global\long\def\cdp{\idp}

\global\long\def\das{}

\global\long\def\id{\mathbb{I}}

\global\long\def\idarg#1#2{\id\left\{  #1,#2\right\}  }

\global\long\def\iid{\stackrel{\text{iid}}{\sim}}

\global\long\def\bzero{\vt 0}

\global\long\def\bone{\mathbf{1}}

\global\long\def\boldm{\boldsymbol{m}}

\global\long\def\bff{\vt f}

\global\long\def\bx{\boldsymbol{x}}

\global\long\def\bl{\boldsymbol{l}}

\global\long\def\be{\boldsymbol{e}}

\global\long\def\bu{\boldsymbol{u}}

\global\long\def\bo{\boldsymbol{o}}

\global\long\def\bh{\boldsymbol{h}}

\global\long\def\bp{\boldsymbol{p}}

\global\long\def\bq{\boldsymbol{q}}

\global\long\def\bs{\boldsymbol{s}}

\global\long\def\bz{\boldsymbol{z}}

\global\long\def\xnew{y}

\global\long\def\bxnew{\boldsymbol{y}}

\global\long\def\bX{\boldsymbol{X}}

\global\long\def\tbx{\tilde{\bx}}

\global\long\def\by{\boldsymbol{y}}

\global\long\def\bY{\boldsymbol{Y}}

\global\long\def\bZ{\boldsymbol{Z}}

\global\long\def\bU{\boldsymbol{U}}

\global\long\def\bv{\boldsymbol{v}}

\global\long\def\bn{\boldsymbol{n}}

\global\long\def\bV{\boldsymbol{V}}

\global\long\def\bI{\boldsymbol{I}}

\global\long\def\bw{\vt w}

\global\long\def\balpha{\gvt{\alpha}}

\global\long\def\bbeta{\gvt{\beta}}

\global\long\def\bmu{\gvt{\mu}}

\global\long\def\btheta{\boldsymbol{\theta}}

\global\long\def\blambda{\boldsymbol{\lambda}}

\global\long\def\bgamma{\boldsymbol{\gamma}}

\global\long\def\bpsi{\boldsymbol{\psi}}

\global\long\def\bphi{\boldsymbol{\phi}}

\global\long\def\bpi{\boldsymbol{\pi}}

\global\long\def\bomega{\boldsymbol{\omega}}

\global\long\def\bepsilon{\boldsymbol{\epsilon}}

\global\long\def\btau{\boldsymbol{\tau}}

\global\long\def\bxi{\boldsymbol{\xi}}

\global\long\def\realset{\mathbb{R}}

\global\long\def\realn{\realset^{n}}

\global\long\def\integerset{\mathbb{Z}}

\global\long\def\natset{\integerset}

\global\long\def\integer{\integerset}

\global\long\def\natn{\natset^{n}}

\global\long\def\rational{\mathbb{Q}}

\global\long\def\rationaln{\rational^{n}}

\global\long\def\complexset{\mathbb{C}}

\global\long\def\comp{\complexset}

\global\long\def\compl#1{#1^{\text{c}}}

\global\long\def\and{\cap}

\global\long\def\compn{\comp^{n}}

\global\long\def\comb#1#2{\left({#1\atop #2}\right) }

\global\long\def\nchoosek#1#2{\left({#1\atop #2}\right)}

\global\long\def\param{\vt w}

\global\long\def\Param{\Theta}

\global\long\def\meanparam{\gvt{\mu}}

\global\long\def\Meanparam{\mathcal{M}}

\global\long\def\meanmap{\mathbf{m}}

\global\long\def\logpart{A}

\global\long\def\simplex{\Delta}

\global\long\def\simplexn{\simplex^{n}}

\global\long\def\dirproc{\text{DP}}

\global\long\def\ggproc{\text{GG}}

\global\long\def\DP{\text{DP}}

\global\long\def\ndp{\text{nDP}}

\global\long\def\hdp{\text{HDP}}

\global\long\def\gempdf{\text{GEM}}

\global\long\def\rfs{\text{RFS}}

\global\long\def\bernrfs{\text{BernoulliRFS}}

\global\long\def\poissrfs{\text{PoissonRFS}}

\global\long\def\grad{\gradient}
 \global\long\def\gradient{\nabla}

\global\long\def\partdev#1#2{\partialdev{#1}{#2}}
 \global\long\def\partialdev#1#2{\frac{\partial#1}{\partial#2}}

\global\long\def\partddev#1#2{\partialdevdev{#1}{#2}}
 \global\long\def\partialdevdev#1#2{\frac{\partial^{2}#1}{\partial#2\partial#2^{\top}}}

\global\long\def\closure{\text{cl}}

\global\long\def\cpr#1#2{\Pr\left(#1\ |\ #2\right)}

\global\long\def\var{\text{Var}}

\global\long\def\Var#1{\text{Var}\left[#1\right]}

\global\long\def\cov{\text{Cov}}

\global\long\def\Cov#1{\cov\left[ #1 \right]}

\global\long\def\COV#1#2{\underset{#2}{\cov}\left[ #1 \right]}

\global\long\def\corr{\text{Corr}}

\global\long\def\sst{\text{T}}

\global\long\def\SST{\sst}

\global\long\def\ess{\mathbb{E}}

\global\long\def\Ess#1{\ess\left[#1\right]}

\newcommandx\ESS[2][usedefault, addprefix=\global, 1=]{\underset{#2}{\ess}\left[#1\right]}

\global\long\def\fisher{\mathcal{F}}

\global\long\def\bfield{\mathcal{B}}
 \global\long\def\borel{\mathcal{B}}

\global\long\def\bernpdf{\text{Bernoulli}}

\global\long\def\betapdf{\text{Beta}}

\global\long\def\dirpdf{\text{Dir}}

\global\long\def\gammapdf{\text{Gamma}}

\global\long\def\gaussden#1#2{\text{Normal}\left(#1, #2 \right) }

\global\long\def\gauss{\mathbf{N}}

\global\long\def\gausspdf#1#2#3{\text{Normal}\left( #1 \lcabra{#2, #3}\right) }

\global\long\def\multpdf{\text{Mult}}

\global\long\def\poiss{\text{Pois}}

\global\long\def\poissonpdf{\text{Poisson}}

\global\long\def\pgpdf{\text{PG}}

\global\long\def\wshpdf{\text{Wish}}

\global\long\def\iwshpdf{\text{InvWish}}

\global\long\def\nwpdf{\text{NW}}

\global\long\def\niwpdf{\text{NIW}}

\global\long\def\studentpdf{\text{Student}}

\global\long\def\unipdf{\text{Uni}}

\global\long\def\transp#1{\transpose{#1}}
 \global\long\def\transpose#1{#1^{\mathsf{T}}}

\global\long\def\mgt{\succ}

\global\long\def\mge{\succeq}

\global\long\def\idenmat{\mathbf{I}}

\global\long\def\trace{\mathrm{tr}}

\global\long\def\argmax#1{\underset{_{#1}}{\text{argmax}} }

\global\long\def\argmin#1{\underset{_{#1}}{\text{argmin}\ } }

\global\long\def\diag{\text{diag}}

\global\long\def\norm{}

\global\long\def\spn{\text{span}}

\global\long\def\vtspace{\mathcal{V}}

\global\long\def\field{\mathcal{F}}
 \global\long\def\ffield{\mathcal{F}}

\global\long\def\inner#1#2{\left\langle #1,#2\right\rangle }
 \global\long\def\iprod#1#2{\inner{#1}{#2}}

\global\long\def\dprod#1#2{#1 \cdot#2}

\global\long\def\norm#1{\left\Vert #1\right\Vert }

\global\long\def\entro{\mathbb{H}}

\global\long\def\entropy{\mathbb{H}}

\global\long\def\Entro#1{\entro\left[#1\right]}

\global\long\def\Entropy#1{\Entro{#1}}

\global\long\def\mutinfo{\mathbb{I}}

\global\long\def\relH{\mathit{D}}

\global\long\def\reldiv#1#2{\relH\left(#1||#2\right)}

\global\long\def\KL{KL}

\global\long\def\KLdiv#1#2{\KL\left(#1\parallel#2\right)}
 \global\long\def\KLdivergence#1#2{\KL\left(#1\ \parallel\ #2\right)}

\global\long\def\crossH{\mathcal{C}}
 \global\long\def\crossentropy{\mathcal{C}}

\global\long\def\crossHxy#1#2{\crossentropy\left(#1\parallel#2\right)}

\global\long\def\breg{\text{BD}}

\global\long\def\lcabra#1{\left|#1\right.}

\global\long\def\lbra#1{\lcabra{#1}}

\global\long\def\rcabra#1{\left.#1\right|}

\global\long\def\rbra#1{\rcabra{#1}}

\global\long\def\model{\text{KGAN}}

\editor{Unknown}

\title{KGAN: How to Break The Minimax Game in GAN}

\author{\name{T}rung Le\email trung.l@deakin.edu.au \\
 \addr Centre for Pattern Recognition and Data Analytics, Australia\\
\AND \name{T}u Dinh Nguyen \email tu.nguyen@deakin.edu.au\\
\addr Centre for Pattern Recognition and Data Analytics, Australia\\
\AND\name{D}inh Phung \email dinh.phung@deakin.edu.au \\
 \addr Centre for Pattern Recognition and Data Analytics, Australia}

\maketitle
\begin{abstract}
Generative Adversarial Networks (GANs) were intuitively and attractively
explained under the perspective of game theory, wherein two involving
parties are a discriminator and a generator. In this game, the task
of the discriminator is to discriminate the real and generated (i.e.,
fake) data, whilst the task of the generator is to generate the fake
data that maximally confuses the discriminator. In this paper, we
propose a new viewpoint for GANs, which is termed as the \emph{minimizing
general loss} viewpoint. This viewpoint shows a connection between
the general loss of a classification problem regarding a convex loss
function and a $f$-divergence between the true and fake data distributions.
Mathematically, we proposed a setting for the classification problem
of the true and fake data, wherein we can prove that the general loss
of this classification problem is exactly the negative $f$-divergence
for a certain convex function $f$. This allows us to interpret the
problem of learning the generator for dismissing the $f$-divergence
between the true and fake data distributions as that of maximizing
the general loss which is equivalent to the min-max problem in GAN
if the Logistic loss is used in the classification problem. However,
this viewpoint strengthens GANs in two ways. First, it allows us to
employ any convex loss function for the discriminator. Second, it
suggests that rather than limiting ourselves in NN-based discriminators,
we can alternatively utilize other powerful families. Bearing this
viewpoint, we then propose using the kernel-based family for discriminators.
This family has two appealing features: i) a powerful capacity in
classifying non-linear nature data and ii) being convex in the feature
space. Using the convexity of this family, we can further develop
Fenchel duality to equivalently transform the max-min problem to the
max-max dual problem. 
\end{abstract}
\begin{keywords}Generative Adversarial Networks, Generative Model,
Fenchel Duality, Mini-max Problem. \end{keywords}

\section{Introduction\label{sec:Introduction}}

Generative model refers to a model that is capable of generating observable
samples abiding by a given data distribution or mimicking the data
samples drawn from an unknown distribution. It is worth studying because
of the following reasons: i) it helps increase our ability to represent
and manipulate high-dimensional probability distributions; ii) generative
models can be incorporated into reinforcement learning in several
ways; and iii) generative models can be trained with missing data
and can provide predictions on inputs that are missing data \citep{goodfellow_nips17_gan_tutorial}.
\begin{figure}[H]
\begin{centering}
\includegraphics[width=0.9\textwidth]{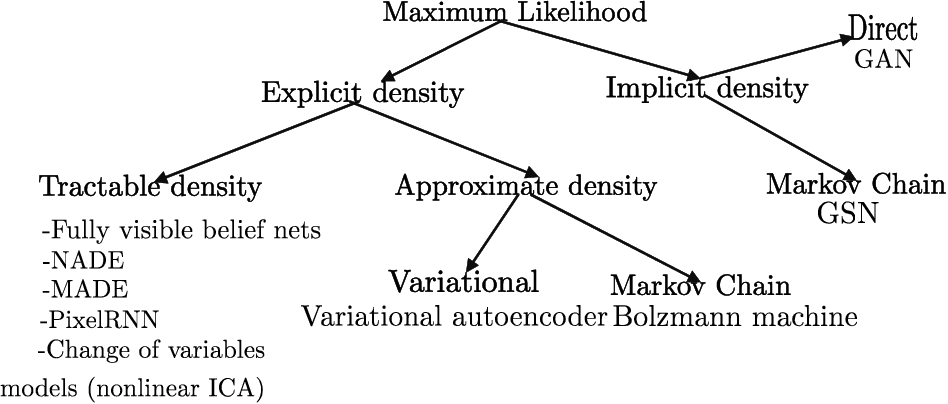}
\par\end{centering}
\caption{The taxonomy of generative models.\label{fig:Taxonomy}}
\end{figure}

The works in generative model can be categorized according to the
taxonomy shown in Figure \ref{fig:Taxonomy} \citep{goodfellow_nips17_gan_tutorial}.
In the left branch of the taxonomic tree, the explicit density node
specifies the models that come with explicit model density function
(i.e., $p_{\text{model}}\left(\bx;\theta\right)$). The maximum likelihood
inference is now straight-forward with an explicit objective function.
The tractability and the precision of inference is totally dependent
on the choice of the density family. This family must be chosen to
be well-presented the true data distribution whilst maintaining the
inference tractable. Under the explicit density node at leftmost,
the tractable density node defines the models whose explicit density
functions are computationally tractable. The well-known models in
this umbrella include fully visible belief nets \citep{Frey:1995:WAP:2998828.2998922},
PixelRNN \citep{oord2016pixel}, Nonlinear ICA \citep{deco1994},
and Real NVP \citep{dinh2016density}. In contrast to the tractable
density node, the approximate density node points out the models that
have explicit density function but are computationally intractable.
The remedy to address this intractability is to approximate the true
density function using either variation method \citep{kingma2013auto,rezende2014stochastic}
or Markov Chain \citep{fahlman1983massively,hinton1984boltzmann}. 

Some generative models can be trained without any model assumption.
These implicit models are pointed out under the umbrella of the implicit
density node. Some of models in this umbrella based on drawing samples
from $p_{\text{model}}\left(\bx;\theta\right)$ formulate a Markov
chain transition operator that must be performed several times to
obtain a sample from the model \citep{BengioT13}. Another existing
state-of-the-art model in this umbrella is Generative Adversarial
Network (GAN) \citep{goodfellow_etal_nips14_gan}. GAN actually introduced
a very novel and powerful way of thinking wherein the generative model
is viewed as a mini-max game consisting of two players (i.e., discriminator
and generator). The discriminator attempts to discriminate the true
data samples against the generated samples, whilst the generator tries
to generate the samples that mimic the true data samples to maximally
challenge the discriminator. The theory behind GAN shows that if the
model converges to the \emph{Nash equilibrium point}, the resulting
generated distribution minimizes its Jensen-Shannon divergence to
the true data distribution \citep{goodfellow_etal_nips14_gan}. The
seminal GAN has really opened a new line of thinking that offers a
foundation for a variety of works \citep{radford2015unsupervised,denton2015deep,ledig2016photo,zhu2016generative,nowozin_etal_nips16_fgan,metz2016unrolled,tu_etal_nips17_d2gan,quan_multi_generator}.
However, because of their mini-max flavor, training GAN(s) is really
challenging. Beside, even if we can perfectly train GAN(s), due to
the nature of the Jensen-Shannon divergence minimization, GAN(s) still
encounter the model collapse issue \citep{theis2015note}. 

In this paper, we first propose to view GAN(s) under another viewpoint,
which is termed as the \emph{minimizing general loss} viewpoint. Intuitively,
since we do not hand in the formulas of both true and generated data
distributions, GAN(s) elegantly invoke a strong discriminator (i.e.,
classifier) to implicitly justify how far these two distributions
are. Concretely, if two distributions are far away, the task of the
discriminator is much easier with a small resulting loss; in contrast,
if they are moving closer, the task of the discriminator becomes harder
with increasingly resulting loss. Eventually, when two distributions
are completely mixed up, the resulting loss of the best discriminator
is maximized, hence we come with the max-min problem, where the inner
minimization is for finding the optimal discriminator given a generator
and the outer maximization is for finding the optimal generator that
maximally makes the optimal discriminator confusing. Mathematically,
we prove that given a convex loss function $\ell\left(\cdot\right)$,
the general loss of the classification for discriminating the true
and fake data is a negative $f$-divergence between the true data
and fake data distributions for a certain convex function $f$. It
follows that we maximize the general loss to minimize the $f$-divergence
between two involving distributions. The viewpoint further explains
why in practice, we can use many loss functions in training GAN while
still gaining good-quality generated samples. Furthermore, the proposed
viewpoint also reveals that we can freely employ any sufficient capacity
family for discriminators instead of limiting ourselves in only NN-based
family. Bearing this observation, we propose using kernel-based discriminators
for classifying the real and fake data. This kernel-based family has
powerful capacity, while being linear convex in the feature space
\citep{Cortes95support-vectornetworks}. This allows us to apply Fenchel
duality to equivalently transform the max-min problem to the max-max
dual problem.

\section{Related Background}

In this section, we present the related background used in our work.
We depart with the introduction of \emph{Fenchel conjugate}, a well-known
notation in convex analysis, followed by the introduction of Fourier
random feature \citep{rahimi_recht_nips07_random_feature} which can
be used to approximate a shift-invariance and positive semi-definite
kernel.

\subsection{Fenchel Conjugate \label{subsec:Fenchel-Conjugate}}

Given a convex function $f:\,S\goto\mathbb{R}$, the Fenchel conjugate
$f^{*}$ of this function is defined as 
\[
f^{*}\left(t\right)=\max_{u\in\text{dom}\left(f\right)}\left(\transp ut-f\left(u\right)\right)
\]

Regarding Fenchel conjugate, we have some following properties:
\begin{enumerate}
\item \textbf{Argmax}: If the function $f$ is strongly convex, the optimal
argument $u^{*}=\text{argmax}_{u}\left(ut-f\left(u\right)\right)$
is exactly $\nabla f^{*}\left(t\right)$. 
\item \textbf{Young inequality}: Given $s,t\in S$, we have the inequality
$f\left(s\right)+f^{*}\left(t\right)\geq st$. The equality occurs
if $t=\nabla f\left(s\right)$.
\item \textbf{Fenchel\textendash Moreau theorem}: If $f$ is convex and
continuous, then the conjugate-ofthe-conjugate (known as the biconjugate)
is the original function: $\left(f^{*}\right)^{*}=f$ which means
that\\
\[
f\left(t\right)=\max_{u\in\text{dom}\left(f^{*}\right)}\left(\transp ut-f^{*}\left(u\right)\right)
\]
\item \textbf{The Legendre transform property}: For strictly convex differentiable
functions, the gradient of the convex conjugate maps a point in the
dual space into the point at which it is the gradient of : $\nabla f^{*}\left(\nabla f\left(t\right)\right)=t$.
\end{enumerate}

\subsection{Fourier Random Feature Representation}

The mapping $\Phi\left(\vectorize x\right)$ above is implicitly defined
and the inner product $\left\langle \Phi\left(\vectorize x\right),\Phi\left(\vectorize x^{\prime}\right)\right\rangle $
is evaluated through a kernel $K\left(\vectorize x,\vectorize x^{\prime}\right)$.
To construct an explicit representation of $\Phi\left(\vectorize x\right)$,
the key idea is to approximate the symmetric and positive semi-definite
(p.s.d) kernel $K\left(\vectorize x,\vectorize x^{\prime}\right)=k\left(\bx-\vectorize x^{\prime}\right)$
with $K\left(\bzero,\bzero\right)=k\left(\bzero\right)=1$ using a
kernel induced by a random finite-dimensional feature map \citep{rahimi_recht_nips07_random_feature}.
The mathematical tool behind this approximation is the Bochner's theorem
\citep{bochner_2016_lectures}, which states that every shift-invariant,
p.s.d kernel $K\left(\vectorize x,\vectorize x^{\prime}\right)$ can
be represented as an inverse Fourier transform of a proper distribution
$p\left(\bomega\right)$ as below:
\begin{equation}
K\left(\vectorize x,\vectorize x^{\prime}\right)=k\left(\bu\right)=\int p\left(\bomega\right)e^{i\bomega^{\top}\bu}d\bomega\label{eq:dist_transform}
\end{equation}
 where $\bu=\vectorize x-\vectorize x^{\prime}$ and $i$ represents
the imaginary unit (i.e., $i^{2}=-1$). In addition, the corresponding
proper distribution $p\left(\boldsymbol{\omega}\right)$ can be recovered
through Fourier transform of kernel function as:
\begin{equation}
p\left(\bomega\right)=\left(\frac{1}{2\pi}\right)^{d}\int k\left(\bu\right)e^{-i\bu^{\top}\bomega}d\bu\label{eq:kernel_transform}
\end{equation}

Popular shift-invariant kernels include Gaussian, Laplacian and Cauchy.
For our work, we employ Gaussian kernel: $K(\bx,\bx')=k\left(\bu\right)=\exp\left[-\frac{1}{2}\bu^{\top}\Sigma\bu\right]$
parameterized by the covariance matrix $\Sigma\in\mathbb{R}^{d\times d}$.
With this choice, substituting into Eq.~(\ref{eq:kernel_transform})
yields a closed-form for the probability distribution $p\left(\bomega\right)$
which is $\mathcal{N}\left(\bzero,\Sigma\right)$.

This suggests a Monte-Carlo approximation to the kernel in Eq.~(\ref{eq:dist_transform}):
\begin{align}
K\left(\vectorize x,\vectorize x^{\prime}\right) & =\mathbb{E}_{\bomega\sim p\left(\bomega\right)}\left[\text{cos}\left(\bomega^{\top}\left(\vectorize x-\vectorize x^{\prime}\right)\right)\right]\approx\sideset{\frac{1}{D}}{_{i=1}^{D}}\sum\left[\cos\left(\bomega_{i}^{\top}\left(\vectorize x-\vectorize x^{\prime}\right)\right)\right]\label{eq:MCMC_approx}
\end{align}
where we have sampled $\boldsymbol{\omega}_{i}\iid\mathcal{N}\left(\boldsymbol{\omega}\gv\bzero,\Sigma\right)$
for $i\in\left\{ 1,2,...,D\right\} $.

Eq.~(\ref{eq:MCMC_approx}) sheds light on the construction of a
$2D$-dimensional random map $\tilde{\Phi}:\mathcal{X}\goto\mathbb{R}^{2D}$:
\begin{align}
\tilde{\Phi}\left(\vectorize x\right) & =\left[\frac{1}{\sqrt{D}}\cos\left(\bomega_{i}^{\top}\vectorize x\right),\frac{1}{\sqrt{D}}\sin\left(\bomega_{i}^{\top}\vectorize x\right)\right]_{i=1}^{D}\label{eq:original_RFF}
\end{align}
resulting in the approximate kernel $\tilde{K}\left(\vectorize x,\vectorize x^{\prime}\right)=\tilde{\Phi}\left(\vectorize x\right)^{\top}\tilde{\Phi}\left(\vectorize x^{\prime}\right)$
that can accurately and efficiently approximate the original kernel:
$\tilde{K}\left(\vectorize x,\vectorize x^{\prime}\right)\approx K\left(\vectorize x,\vectorize x^{\prime}\right)$
\citep{rahimi_recht_nips07_random_feature}.

\subsection{Generative Adversarial Network}

Given a data distribution $\mathbb{P}_{d}$ whose p.d.f is $p_{d}\left(\bx\right)$
where $\bx\in\mathbb{R}^{d}$, the aim of Generative Adversarial Networks
(GAN) \citep{goodfellow_etal_nips14_gan,goodfellow_nips17_gan_tutorial}
is to train a neural-network based generator $G$ such that $G\left(\bz\right)$(s)
fed by $\bz\sim\mathbb{P}_{\bz}$ (i.e., the noise distribution) induce
the generated distribution $\mathbb{P}_{g}$ with the p.d.f $p_{g}\left(\bx\right)$
coinciding the data distribution $\mathbb{P}_{d}$. This is realized
by minimizing the Jensen-Shanon divergence between $\mathbb{P}_{g}$
and $\mathbb{P}_{d}$, which can be equivalently obtained via solving
the following mini-max optimization problem:
\begin{equation}
\text{min}_{G}\text{max}_{D}\left(\mathbb{E}_{\mathbb{P}_{d}}\left[\log\left(D\left(\bx\right)\right)\right]+\mathbb{E}_{\mathbb{P}_{\bz}}\left[\log\left(1-D\left(G\left(\bz\right)\right)\right)\right]\right)\label{eq:minimax_GAN}
\end{equation}
where $D\left(\cdot\right)$ is a neural-network based discriminator
and for a given $\bx$, $D\left(\bx\right)$ specifies the probability
$\bx$ drawn from $\mathbb{P}_{d}$ rather than $\mathbb{P}_{g}$. 

\begin{figure}
\begin{centering}
\includegraphics[width=0.35\textwidth]{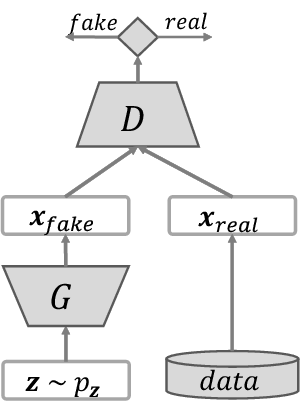}
\par\end{centering}
\caption{The diagram of Generative Adversarial Networks. The generator $G$
produces the generated samples that challenge the discriminator $D$,
while the discriminator tries to differentiate the generated and true
data. \label{fig:GAN_diagram}}

\end{figure}

Under the game theory perspective, GAN can be viewed as a game of
two players: the discriminator $D$ and the generator $G$. The discriminator
tries to discriminate the generated (or fake) data and the real data,
while the generator attempts to make the discriminator confusing by
gradually generating the fake data that break into the real data.
The diagram of GAN is shown in Figure \ref{fig:GAN_diagram}.

Since we do not end up with any formulation for $p_{d}\left(\bx\right)$,
while still being able to generate data from this distribution, GAN(s)
are regarded as a implicit density estimation method. The mysterious
remedy of GANs is to employ a strong discriminator (i.e., classifier)
to implicitly justify the divergence between $\mathbb{P}_{d}$ and
$\mathbb{P}_{g}$. To further clarify this point, we rewrite the optimization
problem in Eq. (\ref{eq:minimax_GAN}) as follows
\begin{gather}
\text{max}_{G}\text{min}_{D}\left(\mathbb{E}_{\mathbb{P}_{d}}\left[\log\left(\frac{1}{D\left(\bx\right)}\right)\right]+\mathbb{E}_{\mathbb{P}_{\bz}}\left[\log\left(\frac{1}{1-D\left(G\left(\bz\right)\right)}\right)\right]\right)\nonumber \\
=\text{max}_{G}\text{min}_{D}\left(\mathbb{E}_{\mathbb{P}_{d}}\left[\log\left(\frac{1}{D\left(\bx\right)}\right)\right]+\mathbb{E}_{\mathbb{P}_{g}}\left[\log\left(\frac{1}{1-D\left(\bx\right)}\right)\right]\right)\label{eq:GAN_loss}
\end{gather}

According the optimization problem in Eq. (\ref{eq:GAN_loss}), given
a generator $G$, we need to train the discriminator $D$ that minimizes
the general logistic loss over the data domain including the real
and fake data. Using the above general loss, we can implicitly estimate
how far $\mathbb{P}_{d}$ and $\mathbb{P}_{g}$ are. In particular,
if $\mathbb{P}_{g}$ is far from $\mathbb{P}_{d}$ then the general
loss is very small, while if $\mathbb{P}_{g}$ is moving closer to
$\mathbb{P}_{d}$ then the general loss increases. In the following
section, we strengthen this by proving that in fact, we can substitute
the logistic loss by any decreasing and convex loss, wherein the the
optimization problem in Eq. (\ref{eq:GAN_loss}) can be equivalently
interpreted as minimizing a certain symmetric $f$-divergence between
$\mathbb{P}_{d}$ and $\mathbb{P}_{g}$.

In addition, the most challenging obstacle in solving the optimization
problem of GAN in Eq. (\ref{eq:minimax_GAN}) is to break its mini-max
flavor. The existing GAN(s) address this problem by alternately updating
the discriminator and generator which cannot accurately solve its
mini-max problem and the rendered solutions might accumulatively diverge
from the optimal one. 

\section{Minimal General Loss Networks}

In this section, we theoretically show the connection between the
problem of discriminating the real and fake data and the problem of
minimizing the distance between $\mathbb{P}_{d}$ and $\mathbb{P}_{g}$.
We start this section with the introduction of the setting for the
classification problem, followed by proving that the general loss
of this classification problem with a certain loss function $\ell\left(\cdot\right)$
is the negative $f$ -divergence of $\mathbb{P}_{d}$ and $\mathbb{P}_{g}$
for some convex function $f$. Finally, we close this section by indicating
some common pairs of $\left(\ell,f\right)$.

\subsection{The Setting of The Classification Problem}

Given two distributions $\mathbb{P}_{d}$ and $\mathbb{P}_{g}$ with
the p.d.f(s) $p_{d}\left(\bx\right)$ and $p_{g}\left(\bx\right)$
respectively, we define the distribution for generating common data
instances as the mixture of two aforementioned distributions as
\[
p\left(\bx\right)=\frac{1}{2}p_{d}\left(\bx\right)+\frac{1}{2}p_{g}\left(\bx\right)\,\text{or}\,\text{\ensuremath{\mathbb{P}\left(\cdot\right)}=\ensuremath{\frac{1}{2}}}\mathbb{P}_{d}\left(\cdot\right)+\frac{1}{2}\mathbb{P}_{g}\left(\cdot\right)
\]

When a data instance $\bx\sim\mathbb{P}$, it is either drawn from
$\mathbb{P}_{d}$ or $\mathbb{P}_{g}$ with the probability $0.5$
for each, we use the following machinery to generate data instance
and label pairs $\left(\bx,y\right)$ where $y\in\left\{ -1,1\right\} $:
\begin{itemize}
\item Randomly draw $\bx\sim\mathbb{P}$.
\item If $\bx$ is really drawn from $\mathbb{P}_{d}$, its label $y$ is
set to $1$. Otherwise, its label $y$ is set to $-1$.
\end{itemize}
Let us denote the joint distribution over $\left(\bx,y\right)$ by
$\mathbb{P}_{x,y}$ whose its p.d.f is $p\left(\bx,y\right)$. It
is evident from our setting that:
\begin{align*}
p\left(\bx\mid y=1\right) & =p_{d}\left(\bx\right)\,\text{and}\,p\left(\bx\mid y=-1\right)=p_{g}\left(\bx\right)\\
\mathbb{P}\left(y=1\right) & =\mathbb{P}\left(y=-1\right)=0.5
\end{align*}

Let $\mathfrak{D}$ be the family of functions with an infinite capacity
that contains discriminators $D\in\mathfrak{D}$, wherein we seek
for the optimal discriminator $D^{*}\in\mathfrak{D}$. To form the
criterion for finding the optimal discriminator, we recruit a decreasing
and convex loss function $\ell:\mathbb{R}\goto\mathbb{R}$. The general
loss w.r.t. a specific discriminator $D$ and the general loss over
the discriminator space are further defined as
\begin{align*}
R_{\ell}\left(D\right) & =\mathbb{E}_{\mathbb{P}_{x,y}}\left[\ell\left(yD\left(\bx\right)\right)\right]\\
R_{\ell}\left(\mathfrak{D}\right) & =\inf_{D\in\mathfrak{D}}R_{\ell}\left(D\right)
\end{align*}

In addition, the optimal discriminator $D^{*}$ is defined as the
discriminator that minimizes the general losses, i.e., $R_{\ell}\left(D^{*}\right)=\inf_{D\in\mathfrak{D}}R_{\ell}\left(D\right)$. 

\subsection{The Relationship between the General Loss and $f$-divergence }

In our setting, we can further derive the general loss over the space
$\mathfrak{D}$ as:
\begin{align*}
R_{\ell}\left(\mathfrak{D}\right) & =\inf_{D\in\mathfrak{D}}R_{\ell}\left(D\right)=\inf_{D\in\mathfrak{D}}\mathbb{E}_{\mathbb{P}_{x,y}}\left[\ell\left(yD\left(\bx\right)\right)\right]=\inf_{D\in\mathfrak{D}}\sum_{y=-1}^{1}\int\ell\left(yD\left(\bx\right)\right)p\left(\bx,y\right)d\bx\\
= & \inf_{D\in\mathfrak{D}}\left\{ \int\ell\left(D\left(\bx\right)\right)p\left(\bx,1\right)d\bx+\int\ell\left(-D\left(\bx\right)\right)p\left(\bx,-1\right)d\bx\right\} \\
= & \frac{1}{2}\inf_{D\in\mathfrak{D}}\left\{ \int\ell\left(D\left(\bx\right)\right)p\left(\bx\mid y=1\right)d\bx+\int\ell\left(-D\left(\bx\right)\right)p\left(\bx\mid y=-1\right)d\bx\right\} \\
= & \frac{1}{2}\inf_{D\in\mathfrak{D}}\left\{ \int\ell\left(D\left(\bx\right)\right)p_{d}\left(\bx\right)d\bx+\int\ell\left(-D\left(\bx\right)\right)p_{g}\left(\bx\right)d\bx\right\} \\
= & \frac{1}{2}\inf_{D\in\mathfrak{D}}\left\{ \int\left[\ell\left(D\left(\bx\right)\right)p_{d}\left(\bx\right)+\ell\left(-D\left(\bx\right)\right)p_{g}\left(\bx\right)\right]d\bx\right\} \\
= & \frac{1}{2}\inf_{D\in\mathfrak{D}}\left\{ \int\left[\ell\left(D\left(\bx\right)\right)\frac{p_{d}\left(\bx\right)}{p_{g}\left(\bx\right)}+\ell\left(-D\left(\bx\right)\right)\right]p_{g}\left(\bx\right)d\bx\right\} 
\end{align*}

Since we assume that the discriminator family $\mathfrak{D}$ has
an infinite capacity, we can proceed the above derivation as follows:
\begin{equation}
R_{\ell}\left(\mathfrak{D}\right)=\frac{1}{2}\int\inf_{\alpha}\left[\ell\left(\alpha\right)\frac{p_{d}\left(\bx\right)}{p_{g}\left(\bx\right)}+\ell\left(-\alpha\right)\right]p_{g}\left(\bx\right)d\bx\label{eq:gen_loss}
\end{equation}

Let us now denote
\begin{equation}
f\left(t\right)=-\inf_{\alpha}\left[\ell\left(\alpha\right)t+\ell\left(-\alpha\right)\right]\label{eq:f_definition}
\end{equation}
, which is a decreasing and convex function, we now plug back this
function to the above formulation to obtain:
\[
R_{\ell}\left(\mathfrak{D}\right)=-\frac{1}{2}\int f\left(\frac{p_{d}\left(\bx\right)}{p_{g}\left(\bx\right)}\right)p_{g}\left(\bx\right)d\bx=-\frac{1}{2}\mathbb{I}_{f}\left(\mathbb{P}_{d}\Vert\mathbb{P}_{g}\right)
\]
 where $\mathbb{I}_{f}\left(\cdot\Vert\cdot\right)$ specifies the
$f$-divergence between two distributions.

It turns out that the general loss is proportional to the negative
$f$-divergence where the convex function $f\left(\cdot\right)$ is
defined as in Eq. (\ref{eq:f_definition}). It also follows that to
minimize $\mathbb{I}_{f}\left(\mathbb{P}_{d}\Vert\mathbb{P}_{g}\right)$,
we can equivalently maximize $R_{\ell}\left(\mathfrak{D}\right)$
and hence come with the following max-min problem:
\[
\sup_{G}\inf_{D}\mathbb{E}_{\mathbb{P}_{x,y}}\left[\ell\left(yD\left(\bx\right)\right)\right]
\]

The above max-min problem also keeps the spirit of GAN(s), which is
the discriminator attempts to classify the real and fake data while
the generator tries to makes the discriminator confusing. From now
on, for the sake of simplification, we replace sup and inf by max
and min, respectively, though the mathematical soundness is lightly
loosen. In particular, we need to tackle the max-min problem:
\[
\max_{G}\min_{D}\mathbb{E}_{\mathbb{P}_{x,y}}\left[\ell\left(yD\left(\bx\right)\right)\right]
\]

It is worth noting that if the loss function $\ell\left(\alpha\right)=\log\left(1+\exp\left(-\alpha\right)\right)$
then the corresponding $f$-divergence is the Jensen-Shannon (JS)
divergence. In Section \ref{subsec:Loss-Divergence}, we will indicate
other loss function and $f$-divergence pairs.

\subsection{Loss Function and $f$-divergence Pairs \label{subsec:Loss-Divergence}}

\subsubsection{0-1 Loss}

This loss has the form $\ell\left(\alpha\right)=\mathbb{I}\left[\alpha\leq0\right]$,
where $\mathbb{I}$ is the indicator function. From Eq. (\ref{eq:gen_loss}),
the optimal discriminator takes the form of $D^{*}\left(\bx\right)=\text{sign}\left(p_{g}\left(\bx\right)-p_{d}\left(\bx\right)\right)$
and the general loss takes the following form:
\begin{align*}
R_{0-1}\left(\mathfrak{D}\right) & =\frac{1}{2}\int\min\left\{ p_{d}\left(\bx\right),p_{g}\left(\bx\right)\right\} d\bx=\frac{1}{2}\int\left[\frac{p_{d}\left(\bx\right)+p_{g}\left(\bx\right)}{2}-\frac{\left|p_{d}\left(\bx\right)-p_{g}\left(\bx\right)\right|}{2}\right]d\bx\\
 & =\frac{1}{2}\left(1-\mathbb{I}_{TV}\left(\mathbb{P}_{d}\Vert\mathbb{P}_{g}\right)\right)
\end{align*}
where $\mathbb{I}_{TV}$ specifies the total variance distance between
two distributions.

\subsubsection{Hinge Loss}

This loss has the form $\ell\left(\alpha\right)=\max\left\{ 0,1-\alpha\right\} $.
From Eq. (\ref{eq:gen_loss}), the optimal discriminator takes the
form of $D^{*}\left(\bx\right)=\text{sign}\left(p_{g}\left(\bx\right)-p_{d}\left(\bx\right)\right)$
and the general loss takes the following form:
\begin{align*}
R_{\text{Hinge}}\left(\mathfrak{D}\right) & =\frac{1}{2}\int2\min\left\{ p_{d}\left(\bx\right),p_{g}\left(\bx\right)\right\} d\bx=\int\left[\frac{p_{d}\left(\bx\right)+p_{g}\left(\bx\right)}{2}-\frac{\left|p_{d}\left(\bx\right)-p_{g}\left(\bx\right)\right|}{2}\right]d\bx\\
 & =1-\mathbb{I}_{TV}\left(\mathbb{P}_{d}\Vert\mathbb{P}_{g}\right)
\end{align*}

\subsubsection{Exponential Loss}

This loss has the form $\ell\left(\alpha\right)=\exp\left(-\alpha\right)$.
From Eq. (\ref{eq:gen_loss}), the optimal discriminator takes the
form of $D^{*}\left(\bx\right)=\frac{1}{2}\log\frac{p_{d}\left(\bx\right)}{p_{g}\left(\bx\right)}$
and the general loss takes the following form:
\begin{align*}
R_{\text{exp}}\left(\mathfrak{D}\right) & =\frac{1}{2}\int2\sqrt{p_{d}\left(\bx\right)p_{g}\left(\bx\right)}d\bx=\frac{1}{2}\left[2-\int\left(\sqrt{p_{d}\left(\bx\right)}-\sqrt{p_{g}\left(\bx\right)}\right)^{2}d\bx\right]\\
 & =1-\mathbb{I}_{\text{Hellinger}}^{2}\left(\mathbb{P}_{d}\Vert\mathbb{P}_{g}\right)
\end{align*}

\subsubsection{Least Square Loss}

This loss has the form $\ell\left(\alpha\right)=\left(1-\alpha\right)^{2}$.
From Eq. (\ref{eq:gen_loss}), the optimal discriminator takes the
form of $D^{*}\left(\bx\right)=\frac{p_{d}\left(\bx\right)-p_{g}\left(\bx\right)}{p_{d}\left(\bx\right)+p_{g}\left(\bx\right)}$
and the general loss takes the following form:
\begin{align*}
R_{\text{sqr}}\left(\mathfrak{D}\right) & =\frac{1}{2}\int\frac{4p_{d}\left(\bx\right)p_{g}\left(\bx\right)}{p_{d}\left(\bx\right)+p_{g}\left(\bx\right)}d\bx=\frac{1}{2}\left[2-\int\frac{\left(p_{d}\left(\bx\right)-p_{g}\left(\bx\right)\right)^{2}}{p_{d}\left(\bx\right)+p_{g}\left(\bx\right)}d\bx\right]\\
 & =1-\mathbb{I}_{\text{f}}\left(\mathbb{P}_{d}\Vert\mathbb{P}_{g}\right)
\end{align*}
where $f\left(t\right)=\frac{-4t}{t+1}$ with $t\geq0$. In addition,
this $f$-divergence is known as the \emph{triangular discrimination
distance}.

\subsubsection{Logistic Loss}

This loss has the form $\ell\left(\alpha\right)=\log$$\left(1+\exp\left(-\alpha\right)\right)$.
From Eq. (\ref{eq:gen_loss}), the optimal discriminator takes the
form of $D^{*}\left(\bx\right)=\log\frac{p_{d}\left(\bx\right)}{p_{g}\left(\bx\right)}$
and the general loss takes the following form:
\begin{align*}
R_{\text{sqr}}\left(\mathfrak{D}\right) & =\frac{1}{2}\int\left[p_{d}\left(\bx\right)\log\frac{p_{d}\left(\bx\right)+p_{g}\left(\bx\right)}{p_{d}\left(\bx\right)}+p_{g}\left(\bx\right)\log\frac{p_{d}\left(\bx\right)+p_{g}\left(\bx\right)}{p_{g}\left(\bx\right)}\right]d\bx\\
 & =\frac{1}{2}\left[2\log2-\mathbb{I}_{\text{KL}}\left(\mathbb{P}_{d}\Vert\frac{\mathbb{P}_{d}+\mathbb{P}_{g}}{2}\right)-\mathbb{I}_{\text{KL}}\left(\mathbb{P}_{g}\Vert\frac{\mathbb{P}_{d}+\mathbb{P}_{g}}{2}\right)\right]\\
 & =\log2-\mathbb{I}_{\text{JS}}\left(\mathbb{P}_{d}\Vert\mathbb{P}_{g}\right)
\end{align*}
where $\mathbb{I}_{\text{JS}}$ specifies the Jensen-Shannon divergence,
which is a $f$-divergence with $f\left(t\right)=-t\log\frac{t+1}{t}-\log\left(t+1\right)$,
$t\geq0$.

\section{Kernelized Generative Adversarial Networks}

\subsection{The Main Idea of KGAN}

Given a p.s.d, symmetric, and shift-invariant kernel $K\left(\dot{\cdot,\cdot}\right)$
with the feature map $\Phi\left(\cdot\right)$, we consider the Reproducing
Kernel Hilbert space (RKHS) $\mathbb{H}_{K}$ of this kernel as the
discriminator family. Therefore each discriminator parameterized by
a vector $\bw\in\mathbb{H}_{K}$ (i.e., $\bw=\sum_{i}\alpha_{i}\Phi\left(\bz_{i}\right)$)
has the following formulation:
\[
D_{\bw}\left(\bx\right)=\transp{\bw}\Phi\left(\bx\right)=\sum_{i}\alpha_{i}K\left(\bz_{i},\bx\right)
\]

To speed up the computation and enable using the backprop in training,
we approximate $K\left(\cdot,\cdot\right)$ using the random feature
kernel $\tilde{K}\left(\cdot,\cdot\right)$ whose random feature map
is $\tilde{\Phi}\left(\cdot\right)$ and hence enforce the discriminator
family to the RKHS $\mathbb{H}_{\tilde{K}}$ of the approximate kernel.
Each discriminator parameterized by a vector $\bw\in\mathbb{H}_{\tilde{K}}$
(i.e., $\bw=\sum_{i}\alpha_{i}\tilde{\Phi}\left(\bz_{i}\right)$)
has the following formulation:
\[
D_{\bw}\left(\bx\right)=\transp{\bw}\tilde{\Phi}\left(\bx\right)=\sum_{i}\alpha_{i}\tilde{K}\left(\bz_{i},\bx\right)
\]

The max-min problem for minimizing the $f$-divergence between two
distributions $\mathbb{P}_{d}$ and $\mathbb{P}_{g}$ is as follows:
\[
\max_{\bpsi}\min_{\bw}\mathbb{E}_{\mathbb{P}_{x,y}}\left[\ell\left(y\transp{\bw}\tilde{\Phi}\left(\bx\right)\right)\right]
\]
 , where we assume that the generator is a NN-based network parameterized
by $\bpsi$. We can further rewrite the above max-min problem as:
\begin{equation}
\max_{\bpsi}\min_{\bw}\left(\mathbb{E}_{\mathbb{P}_{d}}\left[\ell\left(\transp{\bw}\tilde{\Phi}\left(\bx\right)\right)\right]+\mathbb{E}_{\mathbb{P}_{\bz}}\left[\ell\left(-\transp{\bw}\tilde{\Phi}\left(G_{\Psi}\left(\bz\right)\right)\right)\right]\right)\label{eq:kernel_max_min}
\end{equation}

The advantage of the max-min problem in Eq. (\ref{eq:kernel_max_min})
is that we are employing a very powerful family of discriminators,
but each of them is linear in the RKHS $\mathbb{H}_{\tilde{K}}$ which
opens a door for us to employ the Fenchel duality to elegantly transform
the max-min problem to the max-max problem which is much easier to
tame. Moreover, the max-min problem in Eq. (\ref{eq:kernel_max_min})
can be further explained as using the linear models in the RKHS $\mathbb{H}_{\tilde{K}}$
to enforce two push-forward distributions $\mathbb{P}_{d}^{\mathbb{H}_{\tilde{K}}}$
and $\mathbb{P}_{g}^{\mathbb{H}_{\tilde{K}}}$ of $\mathbb{P}_{d}$
and $\mathbb{P}_{g}$ via the transformation $\tilde{\Phi}$ to be
equal. To further clarify this claim, it is always true that $\mathbb{P}_{d}=\mathbb{P}_{g}$
implies $\mathbb{P}_{d}^{\mathbb{H}_{\tilde{K}}}=\mathbb{P}_{g}^{\mathbb{H}_{\tilde{K}}}$,
while the converse statement holds if $\tilde{\Phi}\left(\cdot\right)$
is a bijection. It is very well-known in kernel method that data become
more compacted in the feature space and linear models in this space
are sufficient to well classify data, hence pushing $\mathbb{P}_{g}^{\mathbb{H}_{\tilde{K}}}$
toward $\mathbb{P}_{d}^{\mathbb{H}_{\tilde{K}}}$.

\subsection{The Fenchel Dual Optimization}

Since in reality, we often do not collect enough data, we usually
employ a regulizer $\Omega\left(\bw\right)$ to avoid overfitting.
We now define the following \emph{convex} objective function with
the regulizer $\Omega\left(\bw\right)$ as:

\[
g_{\bpsi}\left(\bw\right)=\Omega\left(\bw\right)+\mathbb{E}_{\mathbb{P}_{d}}\left[\ell\left(\transp{\bw}\tilde{\Phi}\left(\bx\right)\right)\right]+\mathbb{E}_{\mathbb{P}_{\bz}}\left[\ell\left(-\transp{\bw}\tilde{\Phi}\left(G_{\bpsi}\left(\bz\right)\right)\right)\right]
\]
and propose solving the max-min problem: $\max_{\bpsi}\min_{\bw}g_{\bpsi}\left(\bw\right)$.

We first start with $\min_{\bw}\,g_{\bpsi}\left(\bw\right)$ and derive
as follows:{\small{}
\begin{gather}
\min_{\bw}\,g_{\bpsi}\left(\bw\right)=\min_{\bw}\left[\Omega\left(\bw\right)+\mathbb{E}_{\mathbb{P}_{d}}\left[\ell\left(\transp{\bw}\tilde{\Phi}\left(\bx\right)\right)\right]+\mathbb{E}_{\mathbb{P}_{\bz}}\left[\ell\left(-\transp{\bw}\tilde{\Phi}\left(G_{\bpsi}\left(\bz\right)\right)\right)\right]\right]\nonumber \\
=\min_{\bw}\left[\Omega\left(\bw\right)+\mathbb{E}_{\mathbb{P}_{d}}\left[\max_{u_{\bx}}\left[u_{\bx}\transp{\bw}\tilde{\Phi}\left(\bx\right)-\ell^{*}\left(u_{\bx}\right)\right]\right]+\mathbb{E}_{\mathbb{P}_{\bz}}\left[\max_{v_{\bz}}\left[-v_{\bz}\transp{\bw}\tilde{\Phi}\left(G_{\bpsi}\left(\bz\right)\right)-\ell^{*}\left(v_{\bz}\right)\right]\right]\right]\nonumber \\
=\min_{\bw}\max_{\bu,\bv}\left[\Omega\left(\bw\right)+\mathbb{E}_{\mathbb{P}_{d}}\left[u_{\bx}\transp{\bw}\tilde{\Phi}\left(\bx\right)-\ell^{*}\left(u_{\bx}\right)\right]+\mathbb{E}_{\mathbb{P}_{\bz}}\left[-v_{\bz}\transp{\bw}\tilde{\Phi}\left(G_{\bpsi}\left(\bz\right)\right)-\ell^{*}\left(v_{\bz}\right)\right]\right]\nonumber \\
\geq^{(1)}\max_{\bu,\bv}\min_{\bw}\left[\Omega\left(\bw\right)+\mathbb{E}_{\mathbb{P}_{d}}\left[u_{\bx}\transp{\bw}\tilde{\Phi}\left(\bx\right)-\ell^{*}\left(u_{\bx}\right)\right]+\mathbb{E}_{\mathbb{P}_{\bz}}\left[-v_{\bz}\transp{\bw}\tilde{\Phi}\left(G_{\bpsi}\left(\bz\right)\right)-\ell^{*}\left(v_{\bz}\right)\right]\right]\nonumber \\
=-\min_{\bu,\bv}\max_{\bw}\left[-\Omega\left(\bw\right)-\transp{\bw}\left(\mathbb{E}_{\mathbb{P}_{d}}\left[u_{\bx}\tilde{\Phi}\left(\bx\right)\right]-\mathbb{E}_{\mathbb{P}_{\bz}}\left[v_{\bz}\tilde{\Phi}\left(G_{\bpsi}\left(\bz\right)\right)\right]\right)+\mathbb{E}_{\mathbb{P}_{d}}\left[\ell^{*}\left(u_{\bx}\right)\right]+\mathbb{E}_{\mathbb{P}_{z}}\left[\ell^{*}\left(v_{\bz}\right)\right]\right]\nonumber \\
=-\min_{\bu,\bv}\left[\Omega^{*}\left(-\mathbb{E}_{\mathbb{P}_{d}}\left[u_{\bx}\tilde{\Phi}\left(\bx\right)\right]+\mathbb{E}_{\mathbb{P}_{\bz}}\left[v_{\bz}\tilde{\Phi}\left(G_{\bpsi}\left(\bz\right)\right)\right]\right)+\mathbb{E}_{\mathbb{P}_{d}}\left[\ell^{*}\left(u_{\bx}\right)\right]+\mathbb{E}_{\mathbb{P}_{z}}\left[\ell^{*}\left(v_{\bz}\right)\right]\right]\nonumber \\
=\max_{\bu,\bv}\left[-\Omega^{*}\left(-\mathbb{E}_{\mathbb{P}_{d}}\left[u_{\bx}\tilde{\Phi}\left(\bx\right)\right]+\mathbb{E}_{\mathbb{P}_{\bz}}\left[v_{\bz}\tilde{\Phi}\left(G_{\bpsi}\left(\bz\right)\right)\right]\right)-\mathbb{E}_{\mathbb{P}_{d}}\left[\ell^{*}\left(u_{\bx}\right)\right]-\mathbb{E}_{\mathbb{P}_{z}}\left[\ell^{*}\left(v_{\bz}\right)\right]\right]\label{eq:min_max_derivation}
\end{gather}
where $\bu:\mathcal{X}\goto\mathbb{R}$ and $\bv:\mathcal{Z}\goto\mathbb{R}$
with $\bu\left(\bx\right)=u_{\bx},\,\forall\bx$ and $\bv\left(\bz\right)=v_{\bz},\,\forall\bz$.}{\small \par}

Therefore, we achieve the following inequality:
\begin{equation}
\max_{\bpsi}\min_{\bw}g_{\bpsi}\left(\bw\right)\geq\max_{\bpsi}\max_{\bu,\bv}h_{\bpsi}\left(\bu,\bv\right)\label{eq:max_max}
\end{equation}
where we have defined
\[
h_{\bpsi}\left(\bu,\bv\right)=-\Omega^{*}\left(-\mathbb{E}_{\mathbb{P}_{d}}\left[u_{\bx}\tilde{\Phi}\left(\bx\right)\right]+\mathbb{E}_{\mathbb{P}_{\bz}}\left[v_{\bz}\tilde{\Phi}\left(G_{\bpsi}\left(\bz\right)\right)\right]\right)-\mathbb{E}_{\mathbb{P}_{d}}\left[\ell^{*}\left(u_{\bx}\right)\right]-\mathbb{E}_{\mathbb{P}_{z}}\left[\ell^{*}\left(v_{\bz}\right)\right]
\]

The inequality in Eq. (\ref{eq:max_max}) reveals that instead of
solving the max-min problem $\max_{\bpsi}\min_{\bw}g_{\bpsi}\left(\bw\right)$,
we can alternatively solve the max-max problem $\max_{\bpsi}\max_{\bu,\bv}h_{\bpsi}\left(\bu,\bv\right)$,
which allows us to update the variables simultaneously. The inequality
in Eq. (\ref{eq:max_max}) becomes equality if the inequality (1)
in Eq. (\ref{eq:min_max_derivation}) is an equality. In Section \ref{sec:Theory},
we point out some sufficient conditions for this equality.

\subsection{Regularizers}

We now introduce the regulizers that can be used in our $\model$.
The first regulizer mainly consists of the empirical loss on the training
set like the optimization problem in GAN, whilst the second one really
adds a regularization quantity to the empirical loss.

The first regulizer is of the following form 
\[
\Omega\left(\bw\right)=\begin{cases}
0 & \text{if}\,\norm{\bw}\leq C\\
+\infty & \text{otherwise}
\end{cases}
\]

The corresponding Fenchel duality has the following form:
\begin{align*}
\Omega^{*}\left(\btheta\right) & =\max_{\bw}\left(\transp{\btheta}\bw-\Omega\left(\bw\right)\right)=\max_{\norm{\bw}\leq C}\left(\transp{\btheta}\bw\right)=C\max_{\norm{\bw}\leq1}\transp{\btheta}\bw=C\norm{\bw}_{*}
\end{align*}
where $\norm{\cdot}_{*}$ denotes the dual norm of the norm $\norm{\cdot}$. 

The second regulizer is the $\ell_{2}$ norm:
\[
\Omega\left(\bw\right)=\frac{\lambda}{2}\norm{\bw}_{2}^{2}
\]

The corresponding Fenchel duality has the following form:
\[
\Omega^{*}\left(\btheta\right)=\frac{1}{2\lambda}\norm{\theta}_{2}^{2}
\]

\subsection{The Fenchel Conjugate of Loss Function}

\subsubsection{Logistic Loss}

The Logistic loss has the following form
\[
\ell\left(\alpha\right)=\log\left(1+\exp\left(-\alpha\right)\right)
\]
Its Fenchel conjugate is of the following form
\[
\ell^{*}\left(\alpha\right)=\begin{cases}
\alpha\log\alpha+\left(1-\alpha\right)\log\left(1-\alpha\right) & \text{if}\,0\leq\alpha\leq1\\
+\infty & \text{otherwise}
\end{cases}
\]
here we use the convention $0\log0=0$.

\subsubsection{Hinge Loss}

The Hinge loss has the following form
\[
\ell\left(\alpha\right)=\max\left\{ 0,1-\alpha\right\} 
\]

Its Fenchel conjugate is of the following form
\[
\ell^{*}\left(\alpha\right)=\begin{cases}
\alpha & \text{if}\,0\leq\alpha\leq1\\
+\infty & \text{otherwise}
\end{cases}
\]

\subsubsection{Exponential Loss}

The exponential loss has the following form
\[
\ell\left(\alpha\right)=\exp\left(-\alpha\right)
\]

Its Fenchel conjugate is of the following form
\[
\ell^{*}\left(\alpha\right)=\begin{cases}
-\alpha\log\left(-\alpha\right)+\alpha & \text{if}\,\alpha\leq0\\
0 & \text{otherwise}
\end{cases}
\]

\subsubsection{Least Square Loss}

The least square loss has the following form
\[
\ell\left(\alpha\right)=\left(1-\alpha\right)^{2}
\]

Its Fenchel conjugate is of the following form
\[
\ell^{*}\left(\alpha\right)=\frac{\alpha^{2}}{4}+\alpha
\]

\section{Theory Related to KGAN \label{sec:Theory}}

We are further able to prove that $\tilde{\Phi}:\,\mathcal{X}\goto\mathbb{R}^{2D}$
is an one-to-one feature map if $\text{rank}\left\{ \boldsymbol{\be}_{1},...,\boldsymbol{\be}_{D}\right\} =d$
and $\norm{\Sigma^{1/2}}_{F}\text{diam}\left(\mathcal{X}\right)\max_{1\leq i\leq D}\norm{\boldsymbol{\be}_{i}}<2\pi$
where $\text{diam}\left(\mathcal{X}\right)$ denotes the diameter
of the set $\mathcal{X}$ and $\norm{\Sigma^{1/2}}_{F}$ denotes Frobenius
norm of the matrix $\Sigma^{1/2}$. This is stated in the following
theorem.
\begin{thm}
\label{lem:bijection} If $\Sigma$ is a non-singular matrix (i.e.
positive definite matrix), $\norm{\Sigma^{1/2}}_{F}\text{diam}\left(\mathcal{X}\right)\max_{1\leq i\leq D}\norm{\boldsymbol{\be}_{i}}<2\pi$,
and $\text{rank}\left\{ \boldsymbol{\be}_{1},...,\boldsymbol{\be}_{D}\right\} =d$,
$\tilde{\Phi}:\,\mathcal{X}\goto\mathbb{R}^{2D}$ is an one-to-one
feature map.
\end{thm}

We now state the theorem that shows the relationship of two equations:
$\mathbb{P}_{g}\equiv\mathbb{P}_{d}$ and $\mathbb{P}_{g}^{\mathcal{\mathbb{H}_{\tilde{K}}}}\equiv\mathbb{P}_{d}^{\mathcal{\mathbb{H}_{\tilde{K}}}}$.
It is very obvious that $\mathbb{P}_{g}\equiv\mathbb{P}_{d}$ leads
to $\mathbb{P}_{g}^{\mathbb{H}_{\tilde{K}}}\equiv\mathbb{P}_{d}^{\mathcal{\mathbb{H}_{\tilde{K}}}}$.
We then can prove that the converse statement holds if $\tilde{\Phi}\left(\cdot\right)$
is an one-to-one map.
\begin{prop}
\label{thm:converse}If the random feature map $\tilde{\Phi}\left(\cdot\right)$
is an one-to-one map from $\mathcal{X}$ to $\mathbb{R}^{2D}$, $\mathbb{P}_{g}^{\mathcal{\mathbb{H}_{\tilde{K}}}}\equiv\mathbb{P}_{d}^{\mathcal{\mathbb{H}_{\tilde{K}}}}$
implies $\mathbb{P}_{g}\equiv\mathbb{P}_{d}$.
\end{prop}

We now present and prove some sufficient conditions under which the
max-min problem is equivalent the max-max problem. This equivalence
holds when in Eq. (\ref{eq:min_max_derivation}), we obtain the equality:
\begin{gather*}
\min_{\bw}\max_{\bu,\bv}\text{\ensuremath{\tau}}\left(\bw,\bu,\bv\right)=\max_{\bu,\bv}\min_{\bw}\text{\ensuremath{\tau}}\left(\bw,\bu,\bv\right)
\end{gather*}
where $\tau\left(\bw,\bu,\bv\right)=\Omega\left(\bw\right)+\mathbb{E}_{\mathbb{P}_{d}}\left[u_{\bx}\transp{\bw}\tilde{\Phi}\left(\bx\right)-\ell^{*}\left(u_{\bx}\right)\right]+\mathbb{E}_{\mathbb{P}_{\bz}}\left[-v_{\bz}\transp{\bw}\tilde{\Phi}\left(G_{\bpsi}\left(\bz\right)\right)-\ell^{*}\left(v_{\bz}\right)\right]$.

To achieve some sufficient conditions for the equivalence, we use
the theorems in \citep{sion1958general} which for completeness we
present here.
\begin{thm}
\label{thm:general_minimax}Let $M,N$ be any spaces, $\tau$ is a
function over $M\times N$ that is convex-concave like function, i.e.,
$\tau\left(\cdot,\eta\right)$ is a convex function over $M$ for
all $\eta\in N$ and $\tau\left(\mu,\cdot\right)$ is a concave function
over $N$ for all $\mu\in M$. 

i) If $M$ is compact and $\tau\left(\mu,\eta\right)$ is continuous
in $\mu$ for all $\eta\in N$, $\min_{\mu\in M}\max_{\eta\in N}\tau\left(\mu,\eta\right)=\max_{\eta\in N}\min_{\mu\in M}\tau\left(\mu,\eta\right)$.

ii) If $N$ is compact and $\tau\left(\mu,\eta\right)$ is continuous
in $\eta$ for all $\mu\in M$, $\min_{\mu\in M}\max_{\eta\in N}\tau\left(\mu,\eta\right)=\max_{\eta\in N}\min_{\mu\in M}\tau\left(\mu,\eta\right)$.
\end{thm}

Using Theorem \ref{thm:general_minimax}, we arrive some sufficient
conditions for the equivalence of the max-min and the max-max problems
as stated in Theorem \ref{thm:equivalence}.
\begin{thm}
\label{thm:equivalence}The max-min problem is equivalent to the max-max
problem if one of the following statements holds

i) We limit our discriminator family to $\left\{ D_{\bw}:\bw\in\mathcal{W}\right\} $,
where $\mathcal{W}\subset\mathbb{R}^{2D}$ is a compact set (e.g.,
$\mathcal{W}=\mathcal{\bar{B}}_{r}\left(\bw_{0}\right)=\left\{ \bw:\norm{\bw-\bw_{0}}\leq r\right\} $
or $\mathcal{W}=\prod_{i=1}^{2D}\left[a_{i},b_{i}\right]$).

ii) $\mathbb{P}_{\bz}$ is a discrete distribution, e.g., $\mathbb{P}_{\bz}\left(\cdot\right)=\sum_{i=1}^{M}\pi_{i}\delta_{\bz_{i}}\left(\cdot\right)$
where $\delta_{\bz}$ is the atom measure.
\end{thm}

\section{Conclusion\label{sec:Conclusion}}

In this paper, we have proposed a new viewpoint for GANs, termed as
the \emph{minimizing general loss} viewpoint, which points out a connection
between the general loss of a classification problem regarding a convex
loss function and a certain $f$-divergence between the true and fake
data distributions. In particular, we have proposed a setting for
the classification problem of the true and fake data, wherein we can
prove that the general loss of this classification problem is exactly
the negative $f$-divergence for a certain convex function $f$. This
enables us to convert the problem of learning the generator for minimizing
the $f$-divergence between the true and fake data distributions to
that of maximizing the general loss. This viewpoint extends the loss
function used in discriminators to any convex loss function and suggests
us to use kernel-based discriminators. This family has two appealing
features: i) a powerful capacity in classifying non-linear nature
data and ii) being convex in the feature space, which enables the
application of the Fenchel duality to equivalently transform the max-min
problem to the max-max dual problem.

\appendix

\section{All Proofs}

In this appendix, we present all proofs stated in this manuscript.

\textbf{Proof of Theorem \ref{lem:bijection}}

We need to verify that if $\tilde{\Phi}\left(\vectorize x\right)=\tilde{\Phi}\left(\vectorize x'\right)$
then $\vectorize x=\vectorize x'$. We start with
\begin{align*}
0 & =\norm{\tilde{\Phi}\left(\vectorize x\right)-\tilde{\Phi}\left(\vectorize x'\right)}^{2}=\norm{\tilde{\Phi}\left(\vectorize x\right)}^{2}+\norm{\tilde{\Phi}\left(\vectorize x'\right)}^{2}-2\tilde{K}\left(\vectorize x,\vectorize x'\right)=2-2\tilde{K}\left(\vectorize x,\vectorize x'\right)
\end{align*}
It follows that 
\begin{align}
1 & =\tilde{K}\left(\vectorize x,\vectorize x'\right)=\frac{1}{D}\sum_{i=1}^{D}\left(\cos\left(\vectorize u_{i}\right)\cos\left(\vectorize u{}_{i}'\right)+\sin\left(\vectorize u_{i}\right)\sin\left(\vectorize u{}_{i}'\right)\right)\nonumber \\
 & =\frac{1}{D}\sum_{i=1}^{D}\cos\left(\vectorize u_{i}-\vectorize u_{i}'\right)=\frac{1}{D}\sum_{i=1}^{D}\cos\left(\transp{\boldsymbol{\be}_{i}}\Sigma^{1/2}\left(\vectorize x-\vectorize x'\right)\right)\label{eq:cos_1}
\end{align}
where $\vectorize u_{i}=\transp{\boldsymbol{\be}_{i}}\Sigma^{1/2}\vectorize x$
and $\vectorize u_{i}'=\transp{\boldsymbol{\be}_{i}}\Sigma^{1/2}\vectorize x'$.

With noting that $\cos\left(\transp{\boldsymbol{\be}_{i}}\Sigma^{1/2}\left(\vectorize x-\vectorize x'\right)\right)\leq1,\,\forall i$,
from the equality in Eq. (\ref{eq:cos_1}), we gain that $\cos\left(\transp{\boldsymbol{\be}_{i}}\Sigma^{1/2}\left(\vectorize x-\vectorize x'\right)\right)=1,\,\forall i$.
In addition, we have: $\left|\transp{\boldsymbol{\be}_{i}}\Sigma^{1/2}\left(\vectorize x-\vectorize x'\right)\right|\leq\norm{\boldsymbol{\be}_{i}}\norm{\Sigma^{1/2}}_{F}\norm{\vectorize x-\vectorize x'}\leq\norm{\Sigma^{1/2}}_{F}\text{diam}\left(\mathcal{X}\right)\max_{1\leq i\leq D}\norm{\boldsymbol{\be}_{i}}<2\pi$.
It follows that $\transp{\boldsymbol{\be}_{i}}\Sigma^{1/2}\left(\vectorize x-\vectorize x'\right)=0,\,\forall i$.

Since $\text{rank}\left\{ \boldsymbol{\be}_{1},...,\boldsymbol{\be}_{D}\right\} =d$,
we find $d$ linearly independent vectors inside this set (i.e., $\left\{ \boldsymbol{\be}_{1},...,\boldsymbol{\be}_{D}\right\} $).
Without loss of generality, we assume that they are $\boldsymbol{\be}_{1},...,\boldsymbol{\be}_{d}$.
Combining with the fact that $\Sigma$ is not a singular matrix, we
gain $\transp{\boldsymbol{\be}_{1}}\Sigma^{1/2},...,\transp{\boldsymbol{\be}_{d}}\Sigma^{1/2}$
is also linearly independent. It implies that $\transp{\boldsymbol{\be}_{1}}\Sigma^{1/2},...,\transp{\boldsymbol{\be}_{d}}\Sigma^{1/2}$
is a base of $\mathbb{R}^{d}$. Hence, $\vectorize x-\vectorize x'$
can be represented as linear combination of this base which means
\[
\vectorize x-\vectorize x'=\sum_{i=1}^{d}\alpha_{i}\transp{\boldsymbol{\be}_{i}}\Sigma^{1/2}
\]

It follows that
\begin{align*}
\norm{\vectorize x-\vectorize x'}^{2} & =\left\langle \vectorize x-\vectorize x',\sum_{i=1}^{d}\alpha_{i}\transp{\boldsymbol{\be}_{i}}\Sigma^{1/2}\right\rangle =\sum_{i=1}^{d}\alpha_{i}\transp{\boldsymbol{\be}_{i}}\Sigma^{1/2}\left(\vectorize x-\vectorize x'\right)=0
\end{align*}

Therefore, we arrive at $\vectorize x=\vectorize x'$.

\textbf{Proof of Proposition \ref{thm:converse}}

It is trivial from the fact that $\mathbb{P}_{d}$ and $\mathbb{P}_{g}$
are the pushfoward measures of $\mathbb{P}_{d}^{\mathbb{H}_{\tilde{K}}}$
and $\mathbb{P}_{g}^{\mathbb{H}_{\tilde{K}}}$ via the transformation
$\tilde{\Phi}^{-1}$ . 

\textbf{Proof of Theorem \ref{thm:equivalence}}

It is obvious that $\tau\left(\bw,\bu,\bv\right)=\Omega\left(\bw\right)+\mathbb{E}_{\mathbb{P}_{d}}\left[u_{\bx}\transp{\bw}\tilde{\Phi}\left(\bx\right)-\ell^{*}\left(u_{\bx}\right)\right]+\mathbb{E}_{\mathbb{P}_{\bz}}\left[-v_{\bz}\transp{\bw}\tilde{\Phi}\left(G_{\bpsi}\left(\bz\right)\right)-\ell^{*}\left(v_{\bz}\right)\right]$
is a convex-concavelike function since given $\bu,\bv$, $\tau\left(\bw,\bu,\bv\right)$
is a convex function w.r.t $\bw$ and given $\bw$, this function
is a convex function w.r.t $\bu,\bv$ Our task is to reduce to verifying
that either the domain of $\bw$ or that of $\left(\bu,\bv\right)$
is compact.

i) The domain of $\bw$ is $\mathcal{W}$ which is a compact set.
This leads to the conclusion.

ii) Since $\ell^{*}\left(\cdot\right)$ is only finite on the interval
$\left[a,b\right]$, the domain of $\left(\bu,\bv\right)$ has the
form of $\prod_{i=1}^{N}\left[a_{i},b_{i}\right]\times\prod_{j=1}^{M}\left[c_{j},d_{j}\right]$
which is a compact set. We note that in this case, $\bu=\left[u_{1},...,u_{N}\right]$
and $\bv=\left[v_{1},...,v_{M}\right]$ are two vectors.


\begin{thebibliography}{24}
\providecommand{\natexlab}[1]{#1}
\providecommand{\url}[1]{\texttt{#1}}
\expandafter\ifx\csname urlstyle\endcsname\relax
  \providecommand{\doi}[1]{doi: #1}\else
  \providecommand{\doi}{doi: \begingroup \urlstyle{rm}\Url}\fi

\bibitem[Bengio et~al.(2013)Bengio, Thibodeau-Laufer, and Yosinski]{BengioT13}
Y.~Bengio, E.~Thibodeau-Laufer, and J.~Yosinski.
\newblock Deep generative stochastic networks trainable by backprop.
\newblock \emph{CoRR}, 2013.

\bibitem[Bochner(1959)]{bochner_2016_lectures}
S.~Bochner.
\newblock \emph{Lectures on Fourier Integrals}, volume~42.
\newblock Princeton University Press, 1959.

\bibitem[Cortes and Vapnik(1995)]{Cortes95support-vectornetworks}
Corinna Cortes and Vladimir Vapnik.
\newblock Support-vector networks.
\newblock In \emph{Machine Learning}, pages 273--297, 1995.

\bibitem[Deco and Brauer(1995)]{deco1994}
G.~Deco and W.~Brauer.
\newblock Higher order statistical decorrelation without information loss.
\newblock In \emph{Advances in Neural Information Processing Systems 7}, pages
  247--254. 1995.

\bibitem[Denton et~al.(2015)Denton, Chintala, and Fergus]{denton2015deep}
E.~L. Denton, S.~Chintala, and R.~Fergus.
\newblock Deep generative image models using a laplacian pyramid of adversarial
  networks.
\newblock In \emph{Advances in neural information processing systems}, pages
  1486--1494, 2015.

\bibitem[Dinh et~al.(2016)Dinh, Sohl-Dickstein, and Bengio]{dinh2016density}
L.~Dinh, J.~Sohl-Dickstein, and S.~Bengio.
\newblock Density estimation using real nvp.
\newblock \emph{arXiv preprint arXiv:1605.08803}, 2016.

\bibitem[Fahlman et~al.(1983)Fahlman, Hinton, and
  Sejnowski]{fahlman1983massively}
S.~E Fahlman, G.~E Hinton, and T.~J Sejnowski.
\newblock Massively parallel architectures for al: Netl, thistle, and boltzmann
  machines.
\newblock \emph{Proceedings of AAAI-83109}, 113, 1983.

\bibitem[Frey et~al.(1995)Frey, Hinton, and
  Dayan]{Frey:1995:WAP:2998828.2998922}
B.~J. Frey, G.~E. Hinton, and P.~Dayan.
\newblock Does the wake-sleep algorithm produce good density estimators?
\newblock In \emph{Proceedings of the 8th International Conference on Neural
  Information Processing Systems}, NIPS'95, pages 661--667, 1995.

\bibitem[Goodfellow et~al.(2014)Goodfellow, Pouget-Abadie, Mirza, Xu,
  Warde-Farley, Ozair, Courville, and Bengio]{goodfellow_etal_nips14_gan}
I.~Goodfellow, J.~Pouget-Abadie, M.~Mirza, B.~Xu, D.~Warde-Farley, S.~Ozair,
  A.~Courville, and Y.~Bengio.
\newblock Generative adversarial nets.
\newblock In Z.~Ghahramani, M.~Welling, C.~Cortes, N.~D. Lawrence, and K.~Q.
  Weinberger, editors, \emph{Advances in Neural Information Processing Systems
  (NIPS)}, pages 2672--2680. 2014.

\bibitem[Goodfellow(2017)]{goodfellow_nips17_gan_tutorial}
Ian~J. Goodfellow.
\newblock {NIPS} 2016 tutorial: Generative adversarial networks.
\newblock \emph{CoRR}, 2017.
\newblock URL \url{http://arxiv.org/abs/1701.00160}.

\bibitem[Hinton et~al.(1984)Hinton, Sejnowski, and Ackley]{hinton1984boltzmann}
G.~E. Hinton, T.~J. Sejnowski, and D.~H. Ackley.
\newblock \emph{Boltzmann machines: Constraint satisfaction networks that
  learn}.
\newblock Carnegie-Mellon University, Department of Computer Science
  Pittsburgh, PA, 1984.

\bibitem[Hoang et~al.(2017)Hoang, Nguyen, Le, and Phung]{quan_multi_generator}
Quan Hoang, Tu~Dinh Nguyen, Trung Le, and Dinh~Q. Phung.
\newblock Multi-generator generative adversarial nets.
\newblock \emph{CoRR}, abs/1708.02556, 2017.
\newblock URL \url{http://arxiv.org/abs/1708.02556}.

\bibitem[Kingma and Welling(2013)]{kingma2013auto}
Diederik~P Kingma and Max Welling.
\newblock Auto-encoding variational bayes.
\newblock \emph{arXiv preprint arXiv:1312.6114}, 2013.

\bibitem[Ledig et~al.(2016)Ledig, Theis, Husz{\'a}r, Caballero, Cunningham,
  Acosta, Aitken, Tejani, Totz, and Wang]{ledig2016photo}
C.~Ledig, L.~Theis, F.~Husz{\'a}r, J.~Caballero, A.~Cunningham, A.~Acosta,
  A.~Aitken, A.~Tejani, J.~Totz, and Z.~Wang.
\newblock Photo-realistic single image super-resolution using a generative
  adversarial network.
\newblock \emph{arXiv preprint arXiv:1609.04802}, 2016.

\bibitem[Metz et~al.(2016)Metz, Poole, Pfau, and
  Sohl-Dickstein]{metz2016unrolled}
L.~Metz, B.~Poole, D.~Pfau, and J.~Sohl-Dickstein.
\newblock Unrolled generative adversarial networks.
\newblock \emph{arXiv preprint arXiv:1611.02163}, 2016.

\bibitem[Nguyen et~al.(2017)Nguyen, Le, Vu, and Phung]{tu_etal_nips17_d2gan}
Tu~Dinh Nguyen, Trung Le, Hung Vu, and Dinh Phung.
\newblock Dual discriminator generative adversarial nets.
\newblock In \emph{Advances in Neural Information Processing Systems 29
  (NIPS)}, 2017.

\bibitem[Nowozin et~al.(2016)Nowozin, Cseke, and
  Tomioka]{nowozin_etal_nips16_fgan}
S.~Nowozin, B.~Cseke, and R.gan Tomioka.
\newblock f-gan: Training generative neural samplers using variational
  divergence minimization.
\newblock In \emph{Advances in Neural Information Processing Systems (NIPS)},
  pages 271--279, 2016.

\bibitem[Oord et~al.(2016)Oord, Kalchbrenner, and Kavukcuoglu]{oord2016pixel}
A.~v.~den Oord, N.~Kalchbrenner, and K.~Kavukcuoglu.
\newblock Pixel recurrent neural networks.
\newblock \emph{arXiv preprint arXiv:1601.06759}, 2016.

\bibitem[Radford et~al.(2015)Radford, Metz, and
  Chintala]{radford2015unsupervised}
A.~Radford, L.~Metz, and S.~Chintala.
\newblock Unsupervised representation learning with deep convolutional
  generative adversarial networks.
\newblock \emph{arXiv preprint arXiv:1511.06434}, 2015.

\bibitem[Rahimi and Recht(2007)]{rahimi_recht_nips07_random_feature}
A.~Rahimi and B.~Recht.
\newblock Random features for large-scale kernel machines.
\newblock In \emph{Advances in {N}eural {I}nformation {P}rocessing {S}ystems
  (NIPS)}, pages 1177--1184, 2007.

\bibitem[Rezende et~al.(2014)Rezende, Mohamed, and
  Wierstra]{rezende2014stochastic}
D.~J. Rezende, S.~Mohamed, and D.~Wierstra.
\newblock Stochastic backpropagation and approximate inference in deep
  generative models.
\newblock \emph{arXiv preprint arXiv:1401.4082}, 2014.

\bibitem[Sion(1958)]{sion1958general}
Maurice Sion.
\newblock On general minimax theorems.
\newblock \emph{Pacific Journal of mathematics}, 8\penalty0 (1):\penalty0
  171--176, 1958.

\bibitem[Theis et~al.(2015)Theis, Oord, and Bethge]{theis2015note}
L.~Theis, A.~v.~d. Oord, and M.~Bethge.
\newblock A note on the evaluation of generative models.
\newblock \emph{arXiv preprint arXiv:1511.01844}, 2015.

\bibitem[Zhu et~al.(2016)Zhu, Kr{\"a}henb{\"u}hl, Shechtman, and
  Efros]{zhu2016generative}
J.-Y. Zhu, P.~Kr{\"a}henb{\"u}hl, E.~Shechtman, and A.~A. Efros.
\newblock Generative visual manipulation on the natural image manifold.
\newblock In \emph{European Conference on Computer Vision}, pages 597--613.
  Springer, 2016.

\end{thebibliography}

\end{document}